\documentclass[sigconf]{acmart}
\usepackage{graphicx}
\usepackage{float}
\usepackage{subfigure}
\usepackage{amsmath}
\usepackage{multirow}
\usepackage[commandnameprefix=always]{changes}

\begin{document}

\title{Attention-based Dynamic Graph Convolutional Recurrent Neural Network for Traffic Flow Prediction in Highway Transportation}


\author{Tianpu Zhang}
\orcid{0000-0002-7146-8829}
\affiliation{%
  \institution{North China University of Technology}
   \city{Beijing}
  \country{China}
  \postcode{100144}}
\email{zhangtianpu@hotmail.com}

\author{Weilong Ding}
\email{dingweilong@ncut.edu.cn}
\orcid{0000-0002-9982-5488}
\affiliation{%
  \institution{North China University of Technology}
  \city{Beijing}
  \country{China}
  \postcode{100144}
}

\author{Mengda Xing}
\orcid{0000-0001-9838-4687}
\affiliation{%
  \institution{North China University of Technology}
   \city{Beijing}
  \country{China}
  \postcode{100144}}

\begin{abstract}
  As one of the important tools for spatial feature extraction, graph convolution has been applied in a wide range of fields such as traffic flow prediction. However, current popular works of graph convolution cannot guarantee spatio-temporal consistency in a long period. The ignorance of correlational dynamics, convolutional locality and temporal comprehensiveness would limit predictive accuracy. In this paper, a novel \textbf{\underline{A}}ttention-based \textbf{\underline{D}}ynamic \textbf{\underline{G}}raph \textbf{\underline{C}}onvolutional \textbf{\underline{R}}ecurrent \textbf{\underline{N}}eural \textbf{\underline{N}}etwork (ADGCRNN) is proposed to improve traffic flow prediction in highway transportation. 
Three temporal resolutions of data sequence are effectively integrated by self-attention to extract characteristics; multi-dynamic graphs and their weights are dynamically created to compliantly combine the varying characteristics; a dedicated gated kernel emphasizing highly relative nodes is introduced on these complete graphs to reduce overfitting for graph convolution operations.
Experiments on two public datasets show our work better than state-of-the-art baselines, and case studies of a real Web system prove practical benefit in highway transportation. 
\end{abstract}






\maketitle

\section{Introduction}
\begin{figure}
	\begin{minipage}[t]{0.8\linewidth}
		\centering
		\includegraphics[width=2in]{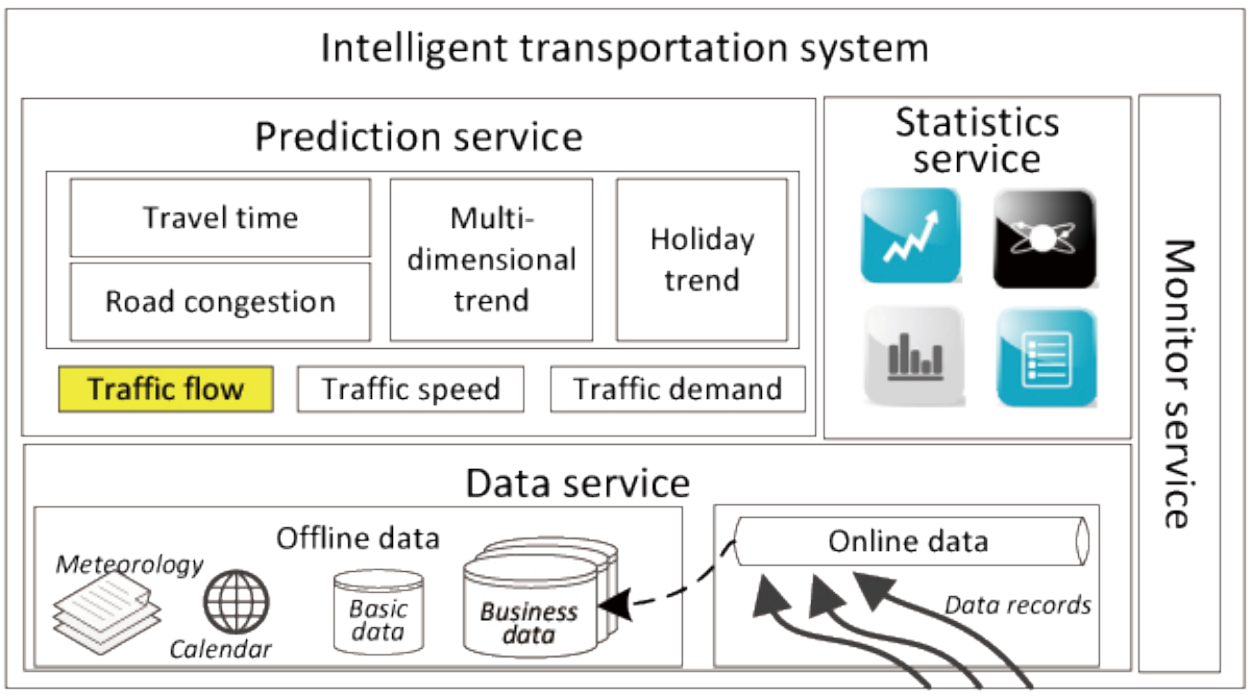}
		\caption{ITS and key components}
		\label{ITS}
	\end{minipage}
	\begin{minipage}[t]{1\linewidth}
		\centering
		\includegraphics[width=3in]{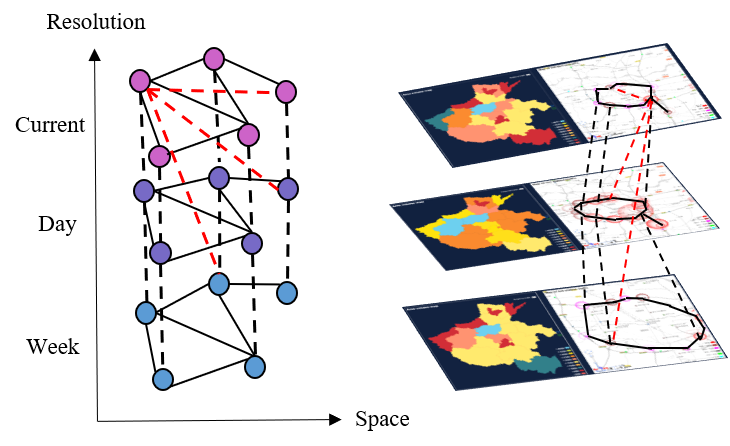}
		\caption{feature of traffic flow at multi-resolutions}
		\label{Multi-resolution spatio-temporal feature}
	\end{minipage}
\end{figure}

Graph convolutional networks (GCN) as one of algorithm of deep learning is widely used in many researches, especially in traffic prediction.
Traffic prediction is one of the most important components of intelligent transportation system (ITS)~\cite{RN3310}. A typical Web-based ITS as Figure \ref{ITS}, is composed of data service, statistics service, prediction service and monitor service. As one of the basic prediction services, traffic flow prediction, focused in this paper, provides significant support for traffic congestion guidance and personal travel planning. In recent years, spatio-temporal features modeling for traffic flow prediction trends to be a hot research topic, especially with the development of GCN. However, spatio-temporal consistency is a major difficulty for graph convolutional networks to guarantee predictive performance in a long-term, because spatial graphs have to be adjusted progressively to fit the feature at that moment. This challenge can be explained in details as follows.

The first is that most of current work lack the consideration of the dynamic relationship between temporal and spatial correlation. The works in \cite{DCRNN,STGCN,STSGCN} only construct a static spatial structure graph with geographic semantics of the structure in highway network, without the consideration of dynamic influence of input features on the road graph. The work ~\cite{T-MGCN} constructs road network topological structure, traffic pattern correlations and area functionality similarities in graphs based on three different semantics respectively. But the weight coefficients among these graphs are all static. Based on road network structure and Dynamic Time Warping (DTW) algorithm,~\cite{STAG-GCN} creates two graph structures in spatial perspectives, but still does not model the dynamics of spatio-temporal features for the weights of those graphs.

The second is that most of works doesn't properly pay attention to the inherent locality of nodes through convolution on the complete graphs. Commonly, the graph created from temporal features is complete graph, and the graph convolution operation on it would aggregate all the nodes of geographic road network. Such information overload inevitably brings predictive overfitting in a long period. For example, the work~\cite{DGCRN} dynamically generates road graphs from temporal features, but its accuracy is limited since the generated adjacency matrix of such graph represents full connectivity. That is, GCN is not easy to aggregate highly correlated information but globally general one of nodes. 

The third is that the long-term dependence at different temporal resolutions is ignored in graph model, which increase the instability and volatility of temporal features. Figure\ref{Multi-resolution spatio-temporal feature} is used as an example, where the left part is schematic and the right correspondingly comes from a real highway ITS developed by us. 
The nodes on the road network naturally construct a spatial graph represented by black solid lines, and can be depicted at different temporal resolution (e.g., current, daily or weekly period). The relationship of a given node at different resolution is showed in black dashed lines. At any resolution, the temporal characteristics of a given node can be affected by that of other nodes at different temporal resolutions, which are represented by red dashed lines. Some current works~\cite{article_1,ST-GDN} only independently consider spatio-temporal features at each resolution through the self-attention mechanism. As a result, predictive performance has to be limited because the models lack comprehensive temporal dependency in a much long period.

In this work, an  Attention-based Dynamic Graph Convolutional Recurrent Neural Network (ADGCRNN) is proposed, which addresses the above drawbacks. The main contributions are summarized as follows.

\begin{itemize}
\item The spatio-temporal relationships of traffic flow sequences at three resolutions are considered to improve predictive accuracy. The temporal resolutions of current, day and week are introduced. By self-attention mechanism, the global spatio-temporal relationships of traffic sequences are effectively integrated with complex spatio-temporal correlations at nodes in a given graph. 

\item A novel dynamic graph cell is introduced to deal with complex spatio-temporal consistency problem. To obtain consistent spatio-temporal features learned from the combination of RNN and GCN at different moments, our model are elaborately designed. To present varying characteristics, the multi-dynamic graphs are dynamically created at different moments based on temporal hidden states in RNN. To compliantly combine those graphs, weights are also dynamically learned. To reduce potential overfitting, GCN locality in complete graphs is emphasize by a ingenious gated kernel.

\item Extensive experiments on real-world datasets and case study in practical project have been conducted. Our work is proved optimal accuracy compared to the state-of-the-art baselines, and show convincing practical benefits distinctly from a real Web-based system.
\end{itemize}

\section{Related Works}
\subsection{Graph Convolution Network for traffic prediction}
Graph convolution network~\cite{GCN}(GCN) has become a popular tool in current work and has been applied in a variety of fields. GCN can be divided into two types. One is to transform spatial domain problem into spectral domain through Fourier transformation to extract spatial features; the other is directly operating convolution in spatial domain. The works~\cite{DualGCN,AdaptiveGCN} are graph convolution models based on the spectral domain and have been used as a baseline in many works. The works~\cite{GAT,DGCNN} are the classical models of second type graph convolution. Here, Graph Attention Network (GAT) adopts attention mechanisms into graph convolution; Deep Graph Convolutional Neural Network (DGCNN) employs SortPolling mechanism to reorder nodes to a meaningful order and then perform the pooling operation.

How to deal with spatio-temporal dependence better is a major problem for traffic flow prediction. Currently, based on a graph convolution network, the majority of works can be divided into two types. One type is to combine RNN and GCN to extract spatio-temporal features. It merges GCN into each cell of RNN, which effectively uses RNN to capture temporal features over a long period of time. With the stacking of RNN's cell, the model can get a greater receptive field and more comprehensive spatio-temporal features. As a typical work of this type,~\cite{AGCRN} fuses GCN with Gated Recurrent Units (GRU)~\cite{GRU} to capture the spatio-temporal dependencies of specific nodes. The work~\cite{DCRNN} applies an encoder-decoder framework based on GRU, merges GCN into each cell of GRU to capture saptio-temporal features, and then finally obtains traffic flow prediction from each cell of the decoder. However, those models consider only one temporal resolution without global temporal dependency, which limits their prediction effects. The other type is to combine convolution neutral network (CNN) and GCN. In this type, CNN is used to extract temporal features and GCN is to extract spatial features. After staking them in turn, the model can obtain spatio-temporal features intensively. The work~\cite{GraphWavenet} belonging to the second type uses temporal convolutional network(TCN)~\cite{TCN}, a variant of CNN, to extract long-time temporal features by dilation convolution with stacking.~\cite{STFGNN} calculates the similarity of traffic sequences by DTW algorithm and stitches multiple graphs into a fusion graph based on similarity to obtain spatio-temporal dependence of the traffic sequences.~\cite{ASTGCN} extracts spatio-temporal features at different temporal resolutions of traffic sequences independently and combines these features in the output layer to obtain final traffic prediction results. Those works above build only the static physical graph, and can't capture dynamical spatio-temporal features varied in a long time . In summary, there is still room to improve traffic flow prediction.

\subsection{Prediction service in Intelligent Transportation System}
Traffic flow, traffic speed and traffic demand are three basic prediction services in Intelligent Transportation System. Traffic flow counts the number of vehicles passing at a given location in a unit time; traffic speed represents average driving speed of all the vehicle across given road segment in a unit time; traffic demand implies the demand for taxis or shared transportation in given areas. Those services have been well studied by domain technicians, and related business functions are adopted progressively in ITS nowadays. For the business simulation, MATSim (multi-agent transport simulation framework)~\cite{MATSim} is a service to model traffic conditions in large-scale urban environment. Based on MATSim,the works ~\cite{RN3163, RN2908} import traffic demand services for shared transportation, such as ridepooling, ride-hailing or taxis. CO-STAR~\cite{CO-STAR} is a dedicated service for highway ITS, but it is for short-term traffic flow prediction in a period of 5 minute. Our work in this paper is long-term (i.e., daily) prediction for traffic flow, and has not been fully optimized compared with the well-studied short-term one in practical ITS currently.

\section{Methodology}
\subsection{Problem Definition}
\begin{figure}
  \centering
  \includegraphics[width=.5\textwidth]{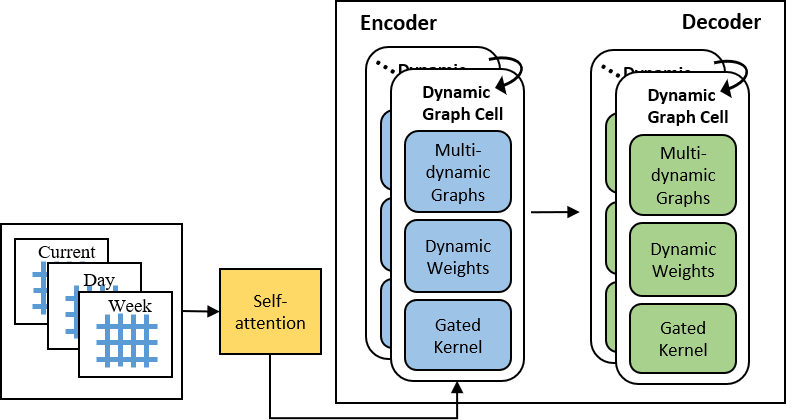}
  \caption{The overview of ADGCRNN} 
  \label{The architecture of SAGCRNN} 
\end{figure}

In this work, the highway graph is represented by an undirected graph $G=(V,E,A)$, where $V$ is the set of node with the number $N$ and the $E$ is the set of edges indicating the connectivity of nodes' pairs. $A\in{\mathbb{R}^{N*N}}$ represents the adjacency matrix of highway graph $G$. $A_{ij}$ represents the connectivity between node $v_i$ and $v_j$, $v_i,v_j\in{V}$.  $A_{ij}=1$ means nodes $v_i$ and $v_j$ are directly connected; $A_{ij}=0$ implies two nodes are not. In this work, traffic flow is the only focus, and the signal matrix of graph $G$ at any time $t$ is $X^{t}\in{\mathbb{R}^N}$ in a frequency of $p$. To obtain a comprehensive view of the spatio-temporal relationships of traffic, we collected the signal matrix of graph at three different resolutions, namely current resolution, day resolution and week resolution. The signal matrix of graph $G$ at time $t$ of different resolutions are $X^t_{current}=X^{t}$, $X^t_{day}=X^{t-p}$ and $X^t_{week}=X^{t-7*p}$. Given a highway graph $G$ and a historical time steps $S$, our prediction problem is to learn a function $\mathbb{F}$ that can predict the signal matrix of graph at $T$ time steps in the future. It can be expressed as the formula~\eqref{proplem defination}. The overview of our work is showed in figure~\ref{The architecture of SAGCRNN}. As the input of the self-attention layer, the signal matrices at different resolutions are combined. Then, the output is pushed into the encoder-decoder module constructed by a dynamic graph cell to capture spatio-temporal features.

\begin{equation}
\{X^{(t+1):(t+T)}\}=\mathbb{F}_G\{X_{current}^{(t-S):t},X_{day}^{(t+1):(t+S)},X_{week}^{(t+1):(t+S)}\}
\label{proplem defination}
\end{equation}

\subsection{Self-attention Layer}
\begin{figure*}
  \centering
  \includegraphics[width=.85\textwidth]{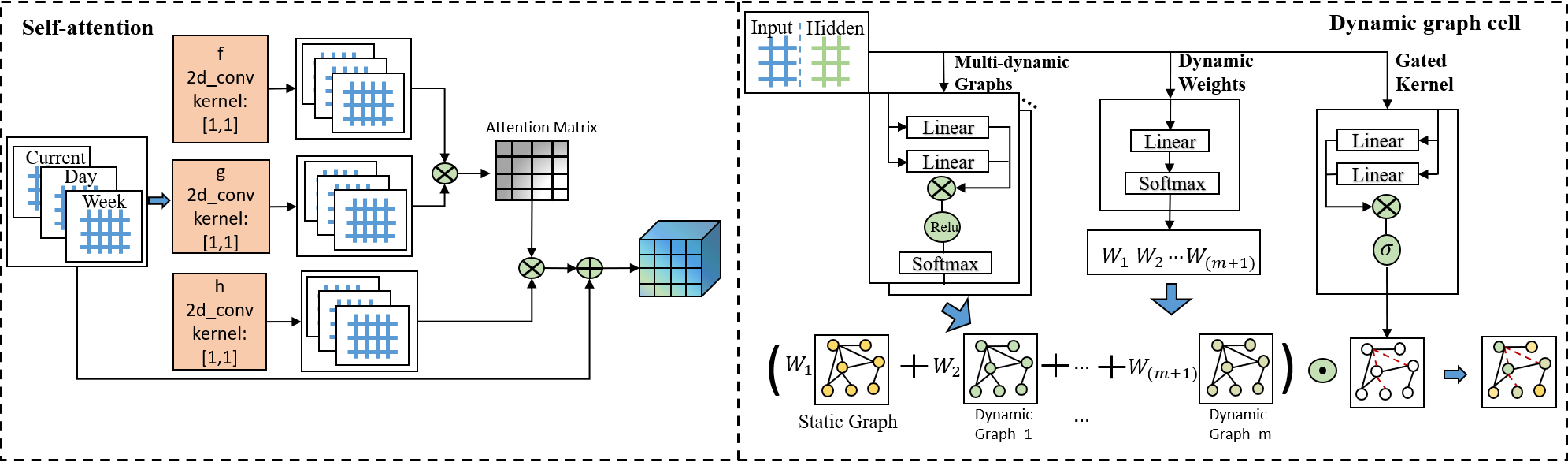}
  \caption{The structure of self-attention and dynamic graph cell}
  \label{The architecutre of self-attention and dynamic graph cell} 
\end{figure*}

Based on the idea that different resolutions of traffic data can bring more comprehensive spatio-temporal features, we construct a self-attention mechanism showed in the left part of figure~\ref{The architecutre of self-attention and dynamic graph cell}. The self-attention mechanism is used to extract spatio-temporal features at three different temporal resolutions, including $X_{current},X_{day},X_{week}$. Unlike ~\cite{ASTGCN} fusing signal matrices of the graph at different resolutions, our work globally considers the interactions among resolutions from the beginning. Our subsequent modules can easily employ the spatio-temporal features over a longer period of time. Therefore, the fusing signal matrices of the graph at different resolutions can be referred to formula~\eqref{merge different resolutions traffic flow}.

\begin{equation}
  \hat{X}^t=X_{current}^t||X_{day}^t||X_{week}^t, \qquad  \hat{X}^t\in{\mathbb{R}^{N*r}}
  \label{merge different resolutions traffic flow}
\end{equation}

Here, notation $||$ denotes feature splicing operation and $\hat{X}^t$ denotes the signal matrix of the graph after fusion three resolutions at time $t$. The signal matrix  $\hat{X}\in{\mathbb{R}^{S*N*r}}$ of traffic flow at $r$ resolutions for $S$ historic steps is input to the self-attention layer, and we project the signal matrix by three 2-D convolutions into the $\mathbb{Q},\mathbb{K},\mathbb{V}$ matrices. The convolutional operation can be expressed by $\mathbb{Q}=\Phi_f*\hat{X}$, $\mathbb{K}=\Phi_g*\hat{X}$, $\mathbb{V}=\Phi_h*\hat{X}$, where $\Phi_f,\Phi_g,\Phi_h\in{\mathbb{R}^{c_{in},c_{out}}}$ are the convolution kernel that are implemented as $1*1$ 2-d convolutions respectively. Further, the self-attention layer is implemented by us through the dot product of matrices, and the formula can be found in ~\eqref{self attention}.

\begin{equation}
  X_{sa}=Att(\mathbb{Q},\mathbb{K},\mathbb{V})=softmax(\mathbb{Q}\mathbb{K}^T)\mathbb{V}+\hat{X}
  \label{self attention}
\end{equation}

Here, notation $X_{sa}\in{\mathbb{R}^{S*N*C_{out}}}$  represents the output of the self-attention layer. The encoded features reflect not only the relationships at the same temporal resolution but also the relationships at different temporal resolutions. As figure~\ref{Multi-resolution spatio-temporal feature}, through 2-d convolution operation, we can aggregate the traffic flow features of the same nodes at different resolutions by the black dashed lines in the figure. The red dashed lines in the figure represents the traffic flow relationship among nodes at different resolutions, which is calculated through matrix dot product operation.

x`

\subsection{Multi-dynamic Graphs}
The intricate spatial relationships of the road network and the time-varying temporal features of traffic flow make it difficult to learn the spatio-temporal features. With those constantly varying temporal features, a static road network topology constructed based on physical semantics is not enough to reflect the dynamic relationships among nodes in a long period. So naturally a topological network, that changes itself at times to fit temporal features of traffic flow, is required. With GRU, our ADGCRNN can dynamically create such topological networks based on the temporal features of the input at different 
time $t$. It is shown in the multi-dynamic graphs section of figure~\ref{The architecutre of self-attention and dynamic graph cell}. 

For any time step $t$, hidden state $H_{t-1}$ of GRU and the output $X_{sa}^t$ of the self-attention layer are concatenated as the input of the dynamic graph cell. It is expressed as formula~\eqref{concatenation_H_X}.

\begin{equation}
  I^t=X_{sa}^t||H^{t-1}
  \label{concatenation_H_X}
\end{equation}

Here, notations $I^t\in{\mathbb{R}^{N*D_{in}}}, H^{t-1}\in{\mathbb{R}^{N*q}}$, $D_{in}$ are the feature dimension after concatenation, and $q$ is the feature dimension of hidden state. Based on input features $I^t$ at time step $t$, the dynamic adjacency matrix $D_g^t\in{\mathbb{R}^{N*N}}$ is created as ~\eqref{Dynamic Graphs}.

\begin{equation}
D_g^t=softmax[Relu(\psi_1(I^t)\psi_2(I^t)^T)]
\label{Dynamic Graphs}
\end{equation}

Here, notations $\psi_1,\psi_2\in{\mathbb{R}^{D_{in}*D_{out}}}$ denote fully connected neural network, which are used to refine temporal features and project them into a new space. The notations $D_{in}$ and $D_{out}$ respectively denote the input feature dimension and output feature dimension of the fully connected network. Further, a multi-head mechanism is used to obtain dynamic adjacency matrices in different semantics and to obtain feature of potential spatial relationships. Specifically, to obtain spatial relationship in more dynamic perspectives, the output features of $\psi_*$ are expanded $m$ folds as $D_{out}*m$. Then, formula \ref{Dynamic Graphs} is applied to create adjacency matrices for all $m$ heads. We eventually merge the dynamic adjacency matrices with the static adjacency matrix to obtain a spatial relationship, as the following formula~\eqref{concatenation_all_graphs}.

\begin{equation}
D_{g\epsilon}^t=D_{g1}^t||...||D_{gm}^t||\hat{A}
\label{concatenation_all_graphs}
\end{equation}

Here, notation $D_{g\epsilon}^t\in{\mathbb{R}^{(m+1)*N*N}}$, where $m$ is a positive integer indicating the number of multi-head. $\hat{A}=\mathbb{D}^{-1}A, \mathbb{D}_{{ii}}=\sum_{j}A_{ij}$ where $i,j\le{N}$ and $\mathbb{D}$ denotes the out-degree diagonal matrix of the adjacency matrix $A$.

\subsection{Dynamic Weight}
Weight coefficients $W^t\in{\mathbb{R}^{(m+1)}}$ are generated for the adjacency matrices, including static adjacency matrix and dynamic adjacency matrices, based on the input features $I^t$ at $t$. It is showed in the dynamic weights part of figure~\ref{The architecutre of self-attention and dynamic graph cell}. The formula to calculate dynamic weights can be found in ~\eqref{dynamic weights}.
\begin{equation}
W^t=softmax(\phi_3(I^t))
\label{dynamic weights}
\end{equation}

Here, notation $\phi_3\in{\mathbb{R}^{D_{in}*(m+1)}}$ denotes the fully connected neural network. $D_{in}$ and $(m+1)$ denote the input feature dimension and output feature dimension of the fully connected network respectively.

\subsection{Gated Kernel}
A gated kernel mechanism is introduced to alleviate information overload problems and to focus locality for convolution operations. As the gated kernel part of figure \ref{The architecutre of self-attention and dynamic graph cell}, an mask matrix $M^t\in{\mathbb{R}^{N*N}}$ is generated based on input $I^t$. This mask matrix $M$ has only two values 0 and 1: $M^t_{ij}=1$ means node $v_j$ has influence on node $v_i$; $M^t_{ij}=0$ has not. Accordingly in the figure\ref{The architecutre of self-attention and dynamic graph cell}, solid black line implies relationship exists between the nodes and red dashed line indicates not after the gated kernel is operated. The formula for the mask matrix $M^t$ is showed in ~\eqref{gated kernel}.

\begin{equation}
M_{ij}^t=\left\{
\begin{aligned}
0,\quad z_{ij}\le{0.5}\\
1 \quad z_{ij}>0.5
\end{aligned}
\right.
,\quad z_{ij}=\sigma[\phi_4(I_{i}^t)\phi_5(I_{j}^t)^T]
\label{gated kernel}
\end{equation}

Here, notation $\phi_4,\phi_5\in{\mathbb{R}^{D_{in}*D_{out}}}$ denotes the respective fully connected neural network, $D_{in}$ and $D_{out}$ denote the input feature dimension and output feature dimension of the fully connected network respectively, and $\sigma$ denotes the sigmoid activation function. The notation $i$ and $j$ are positive integer to indicate the node. As the bottom of the figure \ref{The architecutre of self-attention and dynamic graph cell}, the adjacency matrix, combining static graph with dynamic graphs, is element-wise multiplied by gated kernel $M^t_{ij}$ to make an incomplete graph. Therefore, with this kernel, locality would be emphasized in the graph convolutional operation of dynamic graph cell.

\subsection{Dynamic Graph Cell}
The fusion of the static adjacency matrix with the dynamic adjacency matrix reflects the spatio-temporal correlation among the nodes in different semantics. By graph convolution operation on the fused adjacency matrix, our work can obtain effective spatial features. Here, the diffusion convolution~\cite{DCRNN} is adopted to extract the spatial features among nodes, and the specific formula can be referred to ~\eqref{diffusion convolution}.

\begin{equation}
\begin{aligned}
\hat{D}^t&=W^tD_{g\epsilon}^t \odot M^t\\
\mathbb{X}^t&=\Theta*_G\hat{X}^t=\sum_{k=0}^{K-1}(\hat{D}^t)^k\hat{X}^t
\end{aligned}
\label{diffusion convolution}
\end{equation}

Here, notation $\hat{D}^t\in{\mathbb{R}^{N*N}}$ denotes the dynamic adjacency matrix at time $t$ after  filtering by gated kernel $M^t$. Operation $\odot$ denotes the Hadamard product. In this equation, $\mathbb{X}^t$ denotes the result after the diffusion convolution, and $K$ represents the diffusion coefficient of the diffusion convolution. 

Further, RNN, widely used for serial temporal features, is adopted here, where GRU is employed to improve the capability for long time dependence. As shown in the figure~\ref{The architecture of SAGCRNN}, based on GRU and encoder-decoder framework, we add the previously described multi-dynamic graphs, dynamic weights and gated kernel modules to each of the dynamic graph cells. The graph convolution module is applied to each cell to obtain long time spatio-temporal features, as specified in the formula ~\eqref{GCN_GRU}.

\begin{equation}
\begin{aligned}
r^{t}&=\sigma(\Theta_r*_G(\hat{X}^t||H^{t-1})W_r+b_r)\\
u^{t}&=\sigma(\Theta_u*_G(\hat{X}^t||H^{t-1})W_u+b_u)\\
C^{t}&=tanh(\Theta_C*_G(\hat{X}^{t}||(r^t \odot H^{t-1}))W_C+b_c)\\
H^t&=u^t \odot H^{t-1}+(1-u^t) \odot C^t
\end{aligned}
\label{GCN_GRU}
\end{equation}

Here, notation $||$ indicates the concatenation of features. $W_* \in{\mathbb{R}^{D_{in}*U_{out}}}$, $b_* \in{\mathbb{R}^{U_{out}}}$ are the trainable weight parameters. Operation $\sigma$ denotes the sigmoid activation function and operation $\odot$ denotes the Hadamard product. Operation $*_G$ denotes the diffusion convolution defined in formula~\eqref{diffusion convolution}, $\Theta_r, \Theta_u, \Theta_C$ denotes corresponding graph convolution kernels. We apply the sequence to sequence architecture~\cite{seq_to_seq}, to predict future traffic flow in multiple time steps. To improve the predictive accuracy and efficiency, our ADGCRNN employs the scheduled sampling mechanism~\cite{scheduled_sampling} as follows. In the $i$th training iteration, the probability of $\epsilon_{i}$ is used as the ground truth for the decoder part of the input, and the probability of $1-\epsilon_{i}$ is used as the predicted value for the input. The probability of $\epsilon_{i}$ is gradually reduced to 0 as the count of training iterations $i$ increases. Accordingly, the inconsistent distribution of training and prediction can be eliminated.

\section{Experiment}

\subsection{Datasets}
We evaluate the performance of ADGCRNN on two publicly available traffic datasets, PeMSD4 and PeMSD8 introduced by ~\cite{ASTGCN}. PeMS refers to the Caltrans Performance Measure System, which measures California highway traffic data every 30 seconds.
\begin{itemize}
\item PeMSD4: It represents traffic network dataset for the San Francisco Bay Area. We selected data including a total of 307 nodes, from January to February 2018. 
\item PeMSD8: It represents traffic network dataset for the city of San Bernardino. We selected data including a total of 170 nodes, from July to August 2016.
\item Data Process: We filled missing values in the dataset by linear interpolation and aggregated them every 5 minutes. That is, there are a total of 288 sampling points in a day. The data is then normalised by Z-score method. 
\end{itemize}

\subsection{Baselines}
\begin{itemize}
\item DCRNN~\cite{DCRNN}: Diffusion Convolution Recurrent Neural Network introduced a diffusion convolution and applied GRU model of encoder-decoder.
\item STGCN~\cite{STGCN}: Spatial-temporal Graph Convolution Network combined graph convolution with 1D convolution.
\item GraphWaveNet~\cite{GraphWavenet} Graph WaveNet proposed adaptive adjacency matrix and applied graph convolution with 1D dilated convolution.
\item ASTGCN~\cite{ASTGCN}: Attention Based Spatial Temporal Graph Convolutional Networks obtained spatio-temporal features with different temporal granularity by graph convolution and 2D convolution.
\item AGCRN~\cite{AGCRN}: Adaptive Graph Convolutional Recurrent Network fused graph convolution with RNN networks.
\item STSGCN~\cite{STSGCN}: Spatial-Temporal Synchronous Graph Convolutional Networks designed a spatio-temporal synchronous modelling mechanism to capture local spatio-temporal relationships.
\item STFGNN~\cite{STFGNN}: Spatial-Temporal Fusion Graph Neural Networks constructed a temporal graph by adopting DTW algorithm to obtain local spatio-temporal relationships.
\end{itemize} 

\subsection{Experimental Setting}
All our experiments is implemented and tested on a server with 2 cores Intel Xeon W-2125 CPU, 8 GB RAM, 200 GB storage and GPU NVIDIA GeForce RTX 2080 Ti. ADGCRNN is implemented in the PyTorch framework. 

We divided the dataset into a training set, a validation set and a test set in the ratio of 6:2:2. The parameter daily frequency is set as $p=288$, since traffic flow is calculated in every 5 minutes and a total of 288 periods exist in one day. We use 12 consecutive historical steps to predict 12 consecutive steps in the future, i.e. $S=T=12$. Other hyperparameters setting depends on the performance of validation set. On both test datasets, the best parameters are $C_{out}=3$, $q=32$, $D_{out}=16$, $m=3$, $K=3$. Mean Absolute Error (MAE) and Root Mean Square Error (RMSE) are adopted as metrics here to evaluate accuracy.

\begin{table}[H]
\renewcommand\arraystretch{1.5}
\centering
\caption{Performance on both datasets.}\label{tab_1}
\scalebox{0.9}{
\begin{tabular}{llllll}
\hline
\hline
\multirow{2}*{Model}& Dataset &  \multicolumn{2}{c}{PeMSD4} &\multicolumn{2}{c}{PeMSD8}\\
\cline{2-6}
& Metrics & MAE & RMSE & MAE & RMSE\\
\hline
\multicolumn{2}{c}{STGCN~\cite{STGCN}} & 27.640&42.052&20.870&32.580\\
\hline
\multicolumn{2}{c}{DCRNN~\cite{DCRNN}} & 20.201&32.211&$15.258^*$&$24.166^*$\\
\hline
\multicolumn{2}{c}{Graph WeaveNet~\cite{GraphWavenet}}& 25.672&39.732&19.323&31.229\\
\hline
\multicolumn{2}{c}{ASTGCN~\cite{ASTGCN}}& 25.087&38.669&19.441&30.144\\
\hline
\multicolumn{2}{c}{AGCRN~\cite{AGCRN}}& $19.769^*$&32.221&16.230&25.688\\
\hline
\multicolumn{2}{c}{STSGCN~\cite{STSGCN}}& 21.144&33.540&17.000&26.589\\
\hline
\multicolumn{2}{c}{STFGNN~\cite{STFGNN}}& 19.827&$31.876^*$&16.477&25.943\\
\hline
\hline
\multicolumn{2}{c}{ADGCRNN(ours)}& \textbf{19.008}&\textbf{30.789}&\textbf{14.063}&\textbf{23.348}\\
\hline
\multicolumn{2}{c}{Improvements}& +3.85\%
 & +3.41\% & + 7.83\% & + 3.38\%\\
\end{tabular}}
\end{table}

\subsection{Experimental Result}
The predictive results are showed in table \ref{tab_1}, and the ones marked with * indicate the best among the counterparts. The observed fact demonstrates that our ADGCRNN outperforms others in both MAE and RMSE metrics and its relative improvements are larger than 3.3\%. It proves that our work can learn and fit the spatio-temporal characteristics of traffic flow better by global temporal characteristics at different resolutions. Meanwhile, through dynamically generated adjacency matrices, ADGCRNN can fully employ spatio-temporal features among nodes. All those make our work prominent.

\begin{figure}[htbp]
\centering
\subfigure[MAE in PeMSD8]{
\begin{minipage}[t]{0.5\linewidth}
\centering
\includegraphics[width=1.7in]{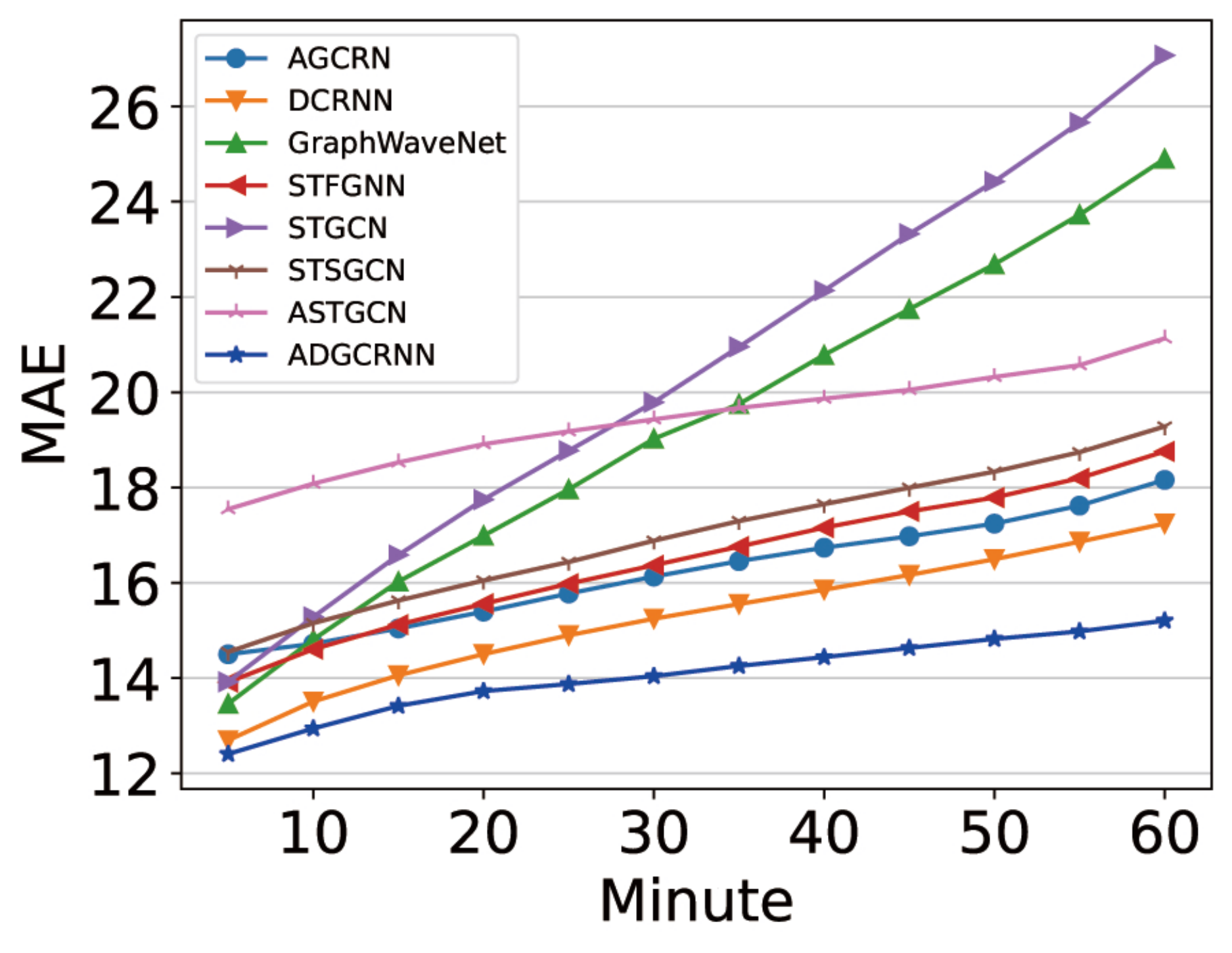}
\end{minipage}%
}%
\subfigure[RMSE in PeMSD8]{
\begin{minipage}[t]{0.5\linewidth}
\centering
\includegraphics[width=1.7in]{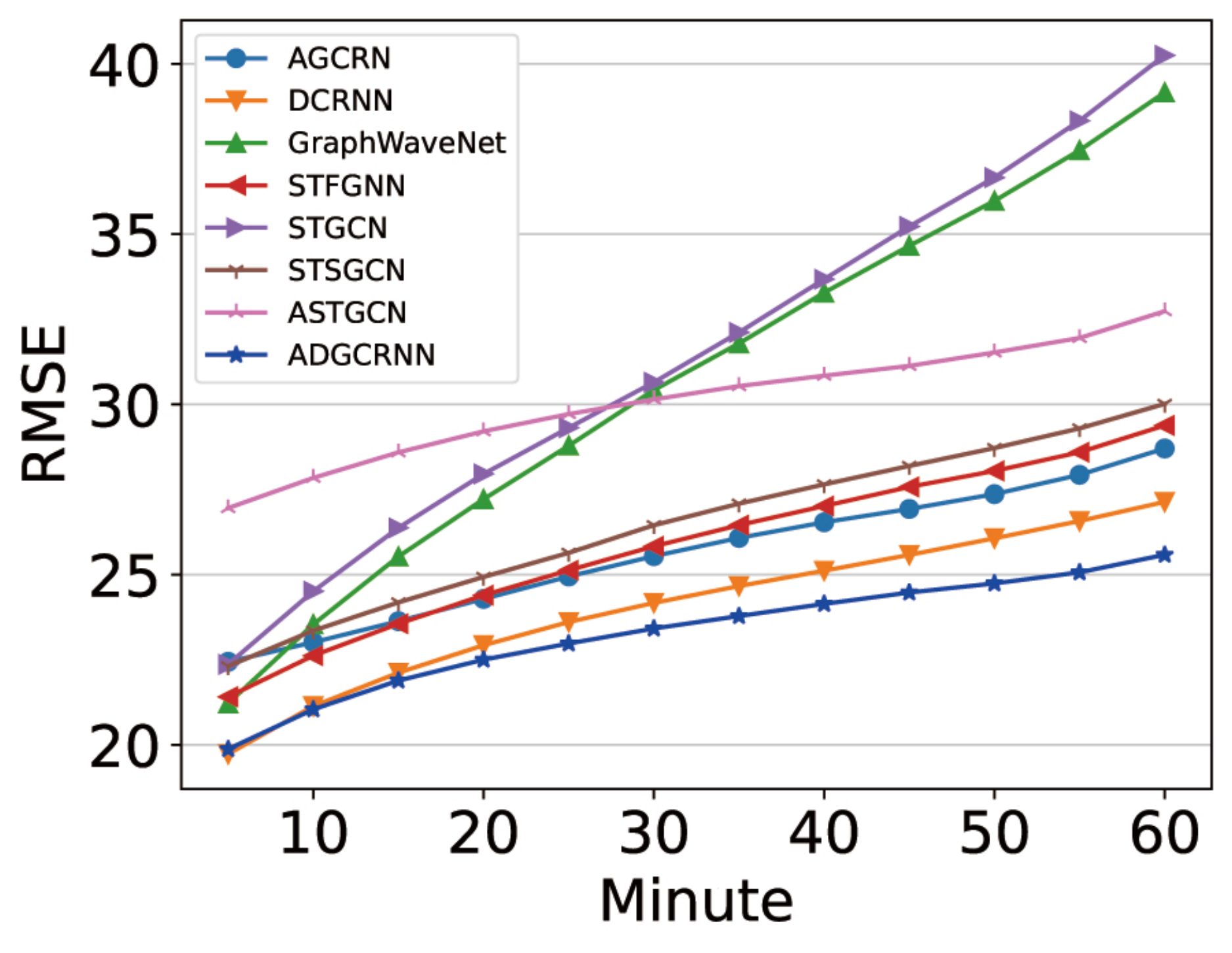}
\end{minipage}%
}%
\qquad
\subfigure[MAE in PeMSD4]{
\begin{minipage}[t]{0.5\linewidth}
\centering
\includegraphics[width=1.7in]{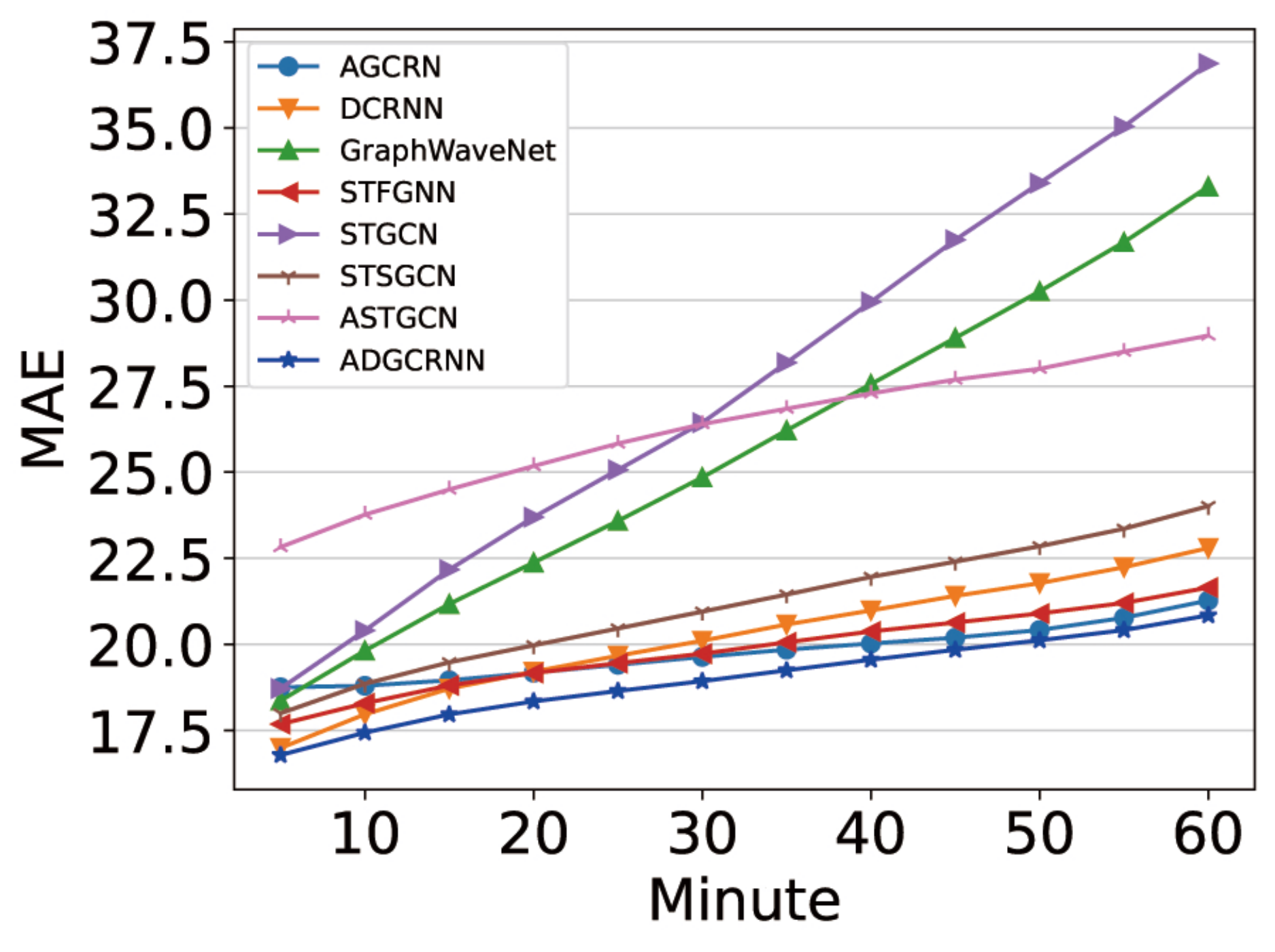}
\end{minipage}%
}%
\subfigure[RMSE in PeMSD4]{
\begin{minipage}[t]{0.5\linewidth}
\centering
\includegraphics[width=1.6in]{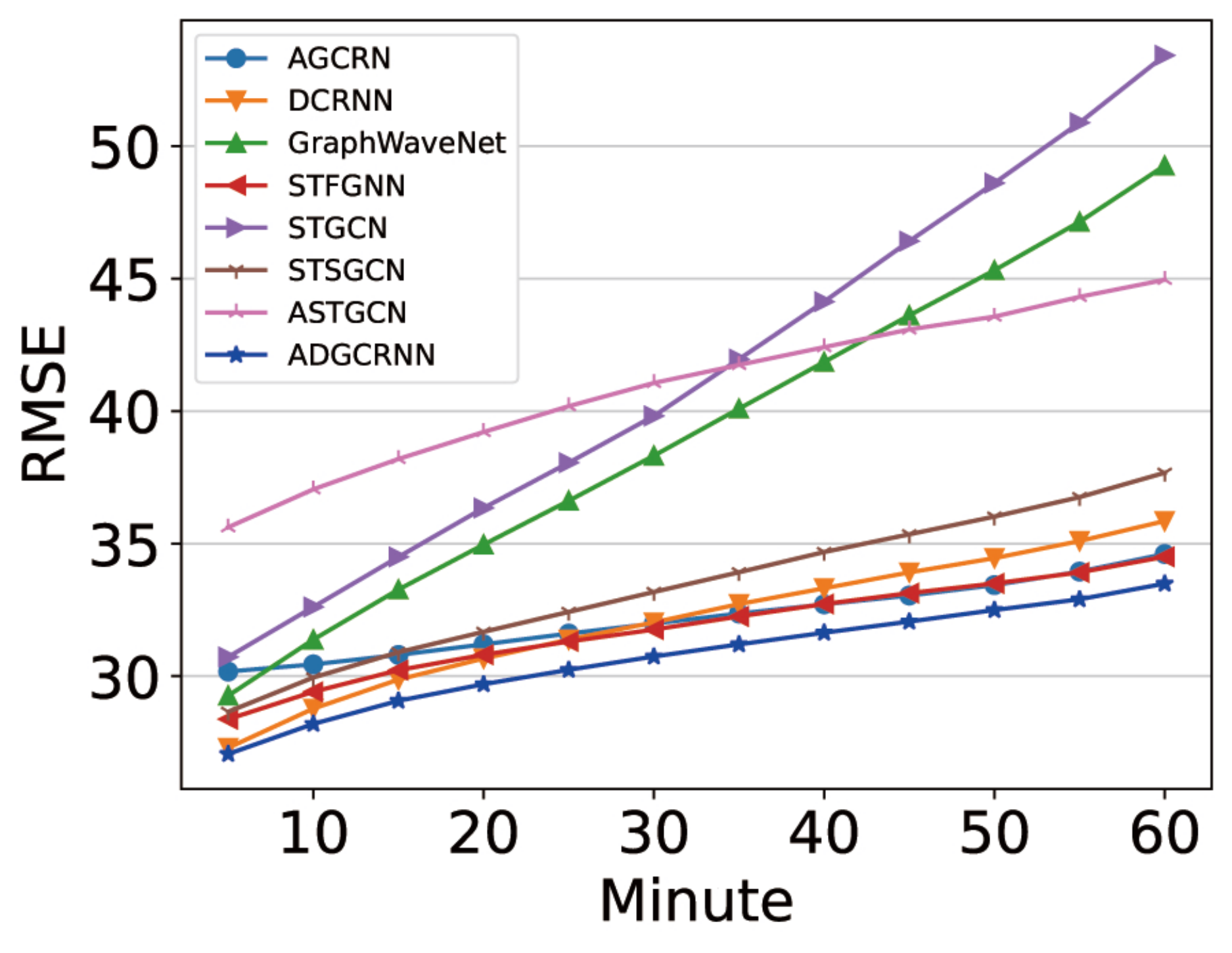}
\end{minipage}%
}
\centering
\caption{Performance at various predictive interval on respective datasets}
\label{each_step}
\end{figure}

Further, we compare the predictive results in details on respective datasets when the prediction interval increases from 5 to 60 minutes. Here, the interval of one prediction step is 5 minutes, when $steps=12$ means that models would predict traffic flow for the next 60 minutes (i.e., 12*5=60). The results are showed in figure~\ref{each_step}. It can be found that the blue line of our ADGCRNN lies below all the other lines, which proves our work outperforms baselines at any prediction interval. There are two reasons to explain that fact. One is that ADGCRNN can learn global temporal features over a longer period of time. The other is that at any interval multi-dynamic graphs generated in ADGCRNN can obtain more comprehensive spatio-temporal features.

\subsection{Ablation Experiment}
To further investigate the effect of different modules on ADGCRNN, three variants are designed and evaluated on PeMSD4 and PeMSD8 datasets. By default, the parameters of ADGCRNN are the same with that of experiment 1. The differences of these variants are described below.

\begin{itemize}
\item $\rm{ADGCRNN_s}$: The model only uses a self-attention mechanism to obtain global temporal relationships of traffic flows at three resolutions. In each dynamic graph cell, it only implements graph convolution with static graph to obtain spatio-temporal features for traffic flow prediction. Its parameters on both datasets are $C_{out}=3$, $q=16$, $K=3$.
\item $\rm{ADGCRNN_{sm}}$: Compared with $\rm{ADGCRNN_{s}}$, this model adds multi-dynamic graphs for dynamic graph cell module to obtain temporal features. Its parameters on both dataset are $C_{out}=3$, $q=32$, $D_{out}=16$, $m=3$, $K=3$. 
\item $\rm{ADGCRNN_{sad}}$: Based on $\rm{ADGCRNN_{sm}}$, this model adds a dynamic weights module for each dynamic graph cell to dynamically adjust relationships among those multiple graphs. Its parameters are the same with that of $\rm{ADGCRNN_{sm}}$. 

\end{itemize}

The results of ablation experiments are shown in the table~\ref{tab_2}. The underlined values indicate the worst performance in this experiment. It can be found that the worst results from variants of ADGCRNN also outperform that of the baselines in experiment 1. It indicates that ADGCRNN is able to better capture the spatio-temporal characteristics of traffic data and obtain more accurate results. Meanwhile, it is clear especially on the PeMSD4 dataset that the predictive results of our work gradually improve with the addition of self-attention mechanism, multi-dynamic graphs module, dynamic weights and gated kernel module. It proves that each of the modules we designed is valuable to better learn the intrinsic spatio-temporal characteristics of traffic flow.

\begin{table}[H]
\renewcommand\arraystretch{1.5}
\centering
\caption{Component analysis of ADGCRNN.}\label{tab_2}
\scalebox{0.95}{
\begin{tabular}{llllll}
\hline
\hline
\multirow{2}*{Model}& Dataset &  \multicolumn{2}{c}{PeMSD4} &\multicolumn{2}{c}{PeMSD8}\\
\cline{2-6}
& Metrics & MAE & RMSE & MAE & RMSE\\
\hline
\multicolumn{2}{c}{$\rm{ADGCRNN_s}$} & \underline{19.767}&\underline{31.784}&14.266&23.885\\
\hline
\multicolumn{2}{c}{$\rm{ADGCRNN_{sm}}$} & 19.449&31.482&14.205&23.579\\
\hline
\multicolumn{2}{c}{$\rm{ADGCRNN_{smd}}$}& 19.258&31.064&\underline{14.309}&\underline{23.973}\\
\hline
\multicolumn{2}{c}{ADGCRNN}& \textbf{19.008}&\textbf{30.789}&\textbf{14.063}&\textbf{23.348}\\
\hline
\end{tabular}}
\end{table}

\subsection{Parametric Analysis of Multi-dynamic Graphs}
\begin{figure}[htbp]
\centering
\subfigure[MAE in PeMSD8]{
\begin{minipage}[t]{0.49\linewidth}
\centering
\includegraphics[width=0.9\textwidth]{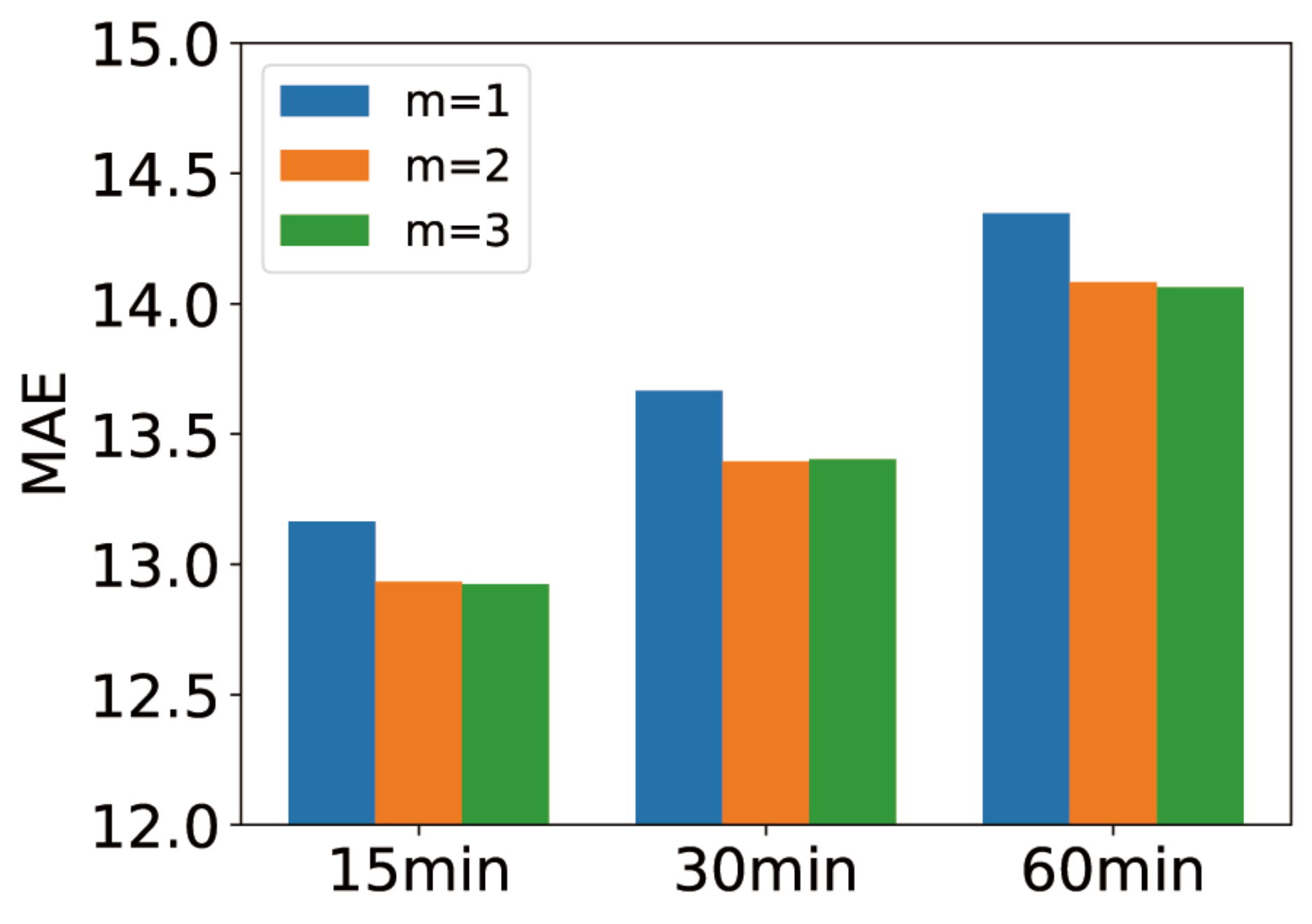}
\end{minipage}%
}%
\subfigure[RMSE in PeMSD8]{
\begin{minipage}[t]{0.49\linewidth}
\centering
\includegraphics[width=0.9\textwidth]{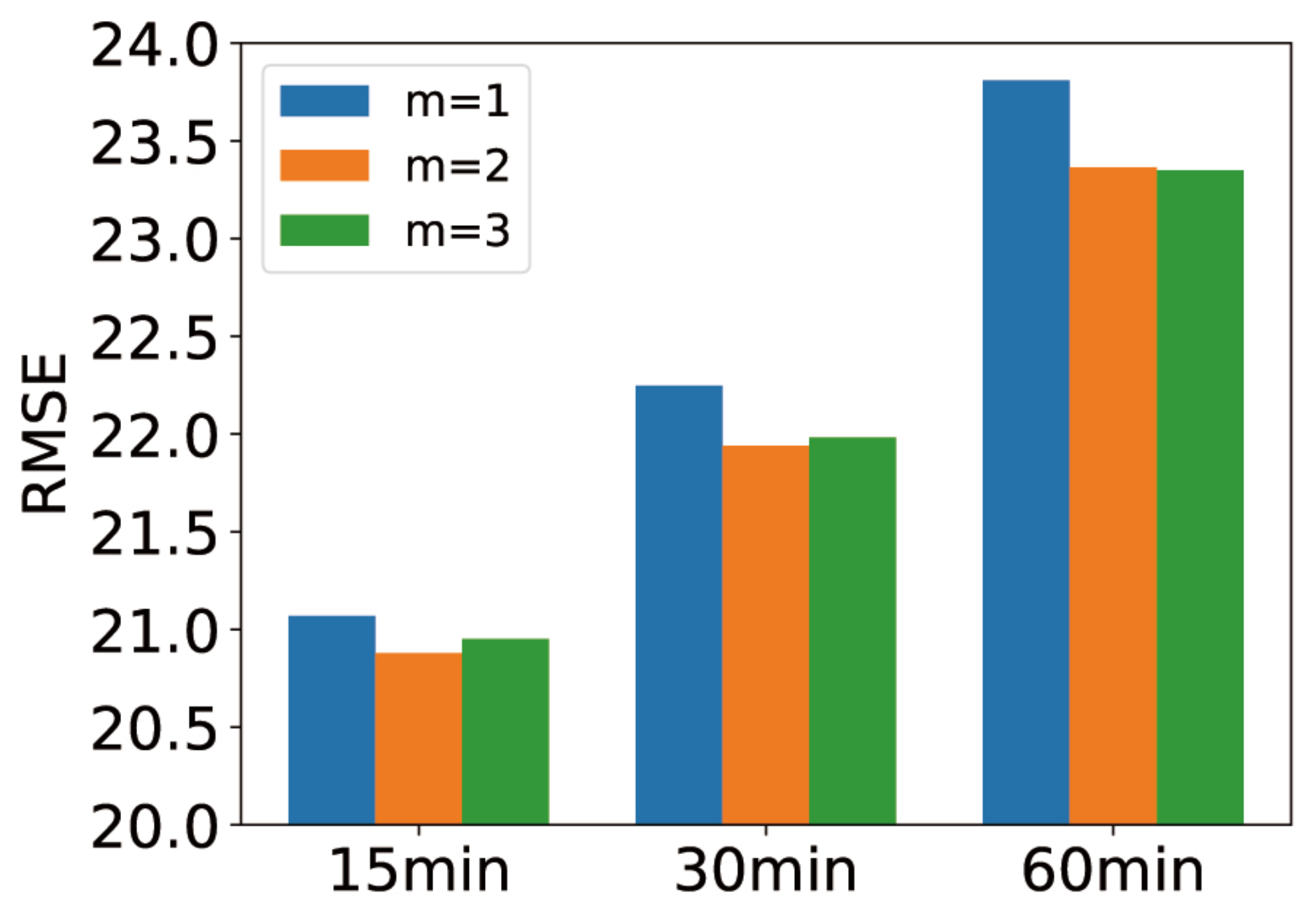}
\end{minipage}%
}%
\qquad
\subfigure[MAE in PeMSD4]{
\begin{minipage}[t]{0.49\linewidth}
\centering
\includegraphics[width=0.9\textwidth]{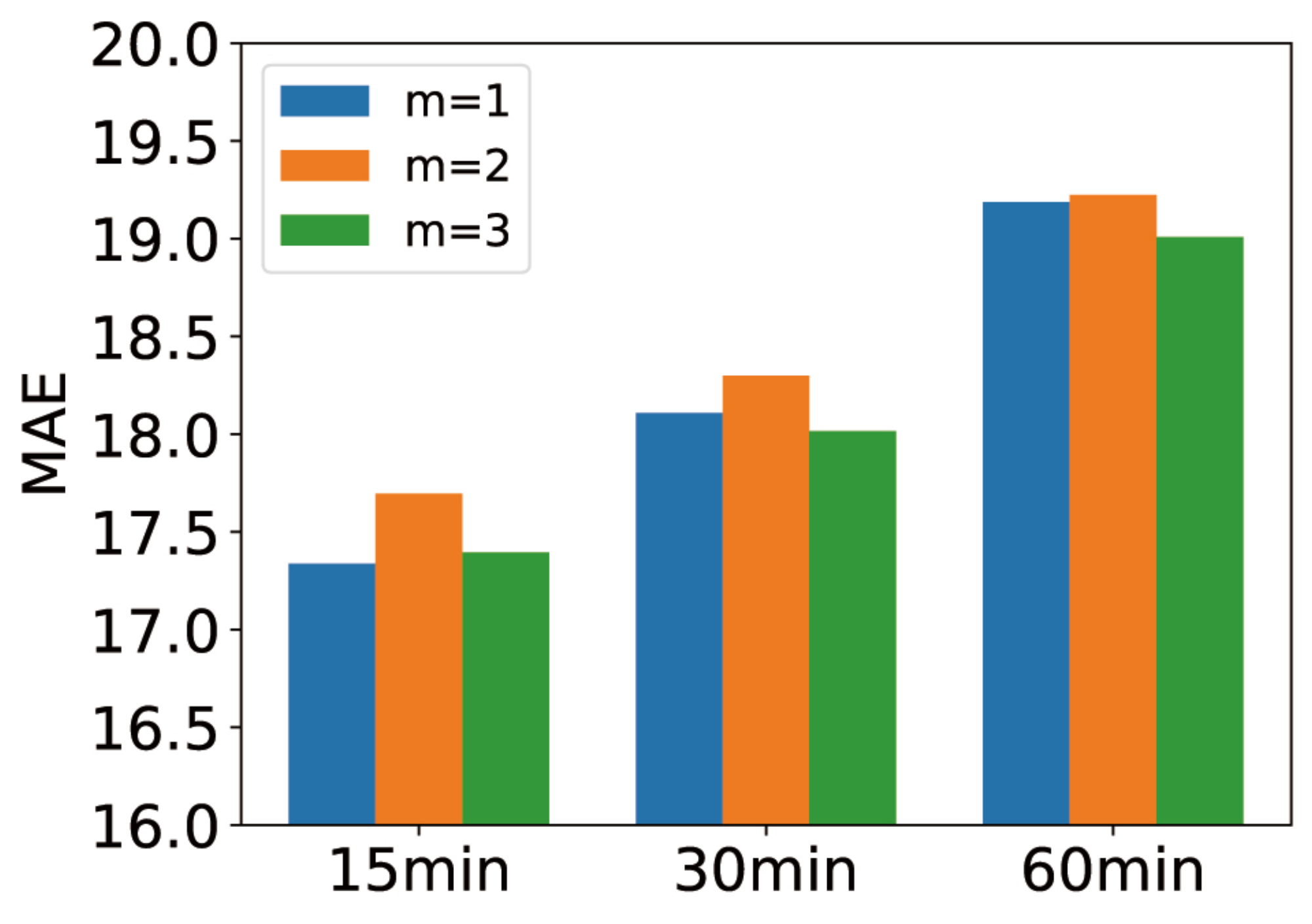}
\end{minipage}%
}%
\subfigure[RMSE in PeMSD4]{
\begin{minipage}[t]{0.49\linewidth}
\centering
\includegraphics[width=0.9\textwidth]{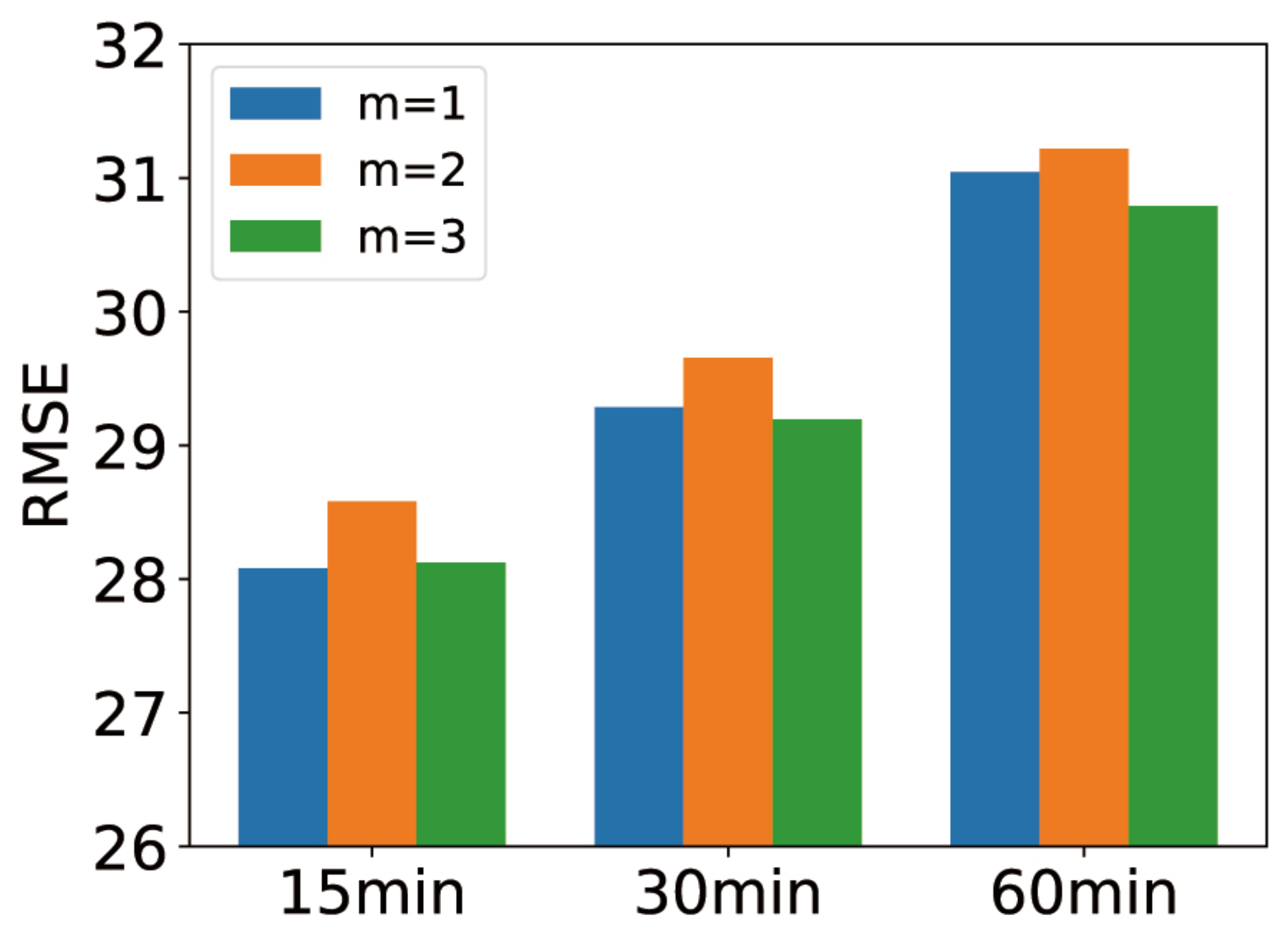}
\end{minipage}%
}
\centering
\caption{Effects on respective datasets with parameters $m$}
\label{different_m}
\end{figure}

It is a key point in our work to capture dynamic and consistent spatio-temporal feature through multi-dynamic graphs. The parameter $m$ decides that how many dynamic graphs with dynamic weights should be generated. Therefore, a experiment is designed here to interpret how this parameter influences the effects in our work. In this experiment, the parameter $m$ is set from 1 to 3, and other parameters are set as $C_{out}=3$, $q=32$, $D_{out}=16$, $m=3$, $K=3$. With different values of $m$, ADGCRNN predicts traffic flow for next 15 minutes, 30 minutes and 60 minutes on two datasets, i.e. $T=3,T=6$ and $T=12$ respectively. The results are referred to figure~\ref{different_m}.

As the figure~\ref{different_m}, the model with $m=3$ in green color has the best performance in the prediction for 30min and 60min. It suggests that more multi-dynamic graphs leads a more comprehensive view of spatio-temporal characteristics and better performance for longer prediction steps. More interesting and counter-intuitive, when $m=3$ our model even achieves the best performance on PeMS8 dataset including less nodes (i.e., 170), compared with that of PeMS4 including more nodes (i.e.,307 ). The sparse nodes in graph network commonly learn fewer spatial relationships, while the result in this experiment implies that more dynamic graphs introduced in ADGCRNN from various perspectives can compensate the feature representation for sparse nodes in graph.

\subsection{Case study}
In this subsection, we show benefits of ADGCRNN in practice. Our work has been applied in \textit{Henan Highway Big Data Analysis System}. This Web-based ITS is built by our team since October 2017 for highway management in \textit{Henan}, the most populous province in China. 
In the system, at 12:00 a.m. everyday, network-wide traffic flow would be predicted for coming days through ADGCRNN. Current 269 toll stations in Henan provincial highway form the nodes of graphs. All the predictive results would be written to Big Data storage. 
Two applications in that system are explained as case studies below. One is situation predictive analysis on holidays, and the other is network-wide monitor in provincial highway.

\begin{figure}
	\begin{minipage}[t]{1\linewidth}
		\centering
		\includegraphics[width=3.5in]{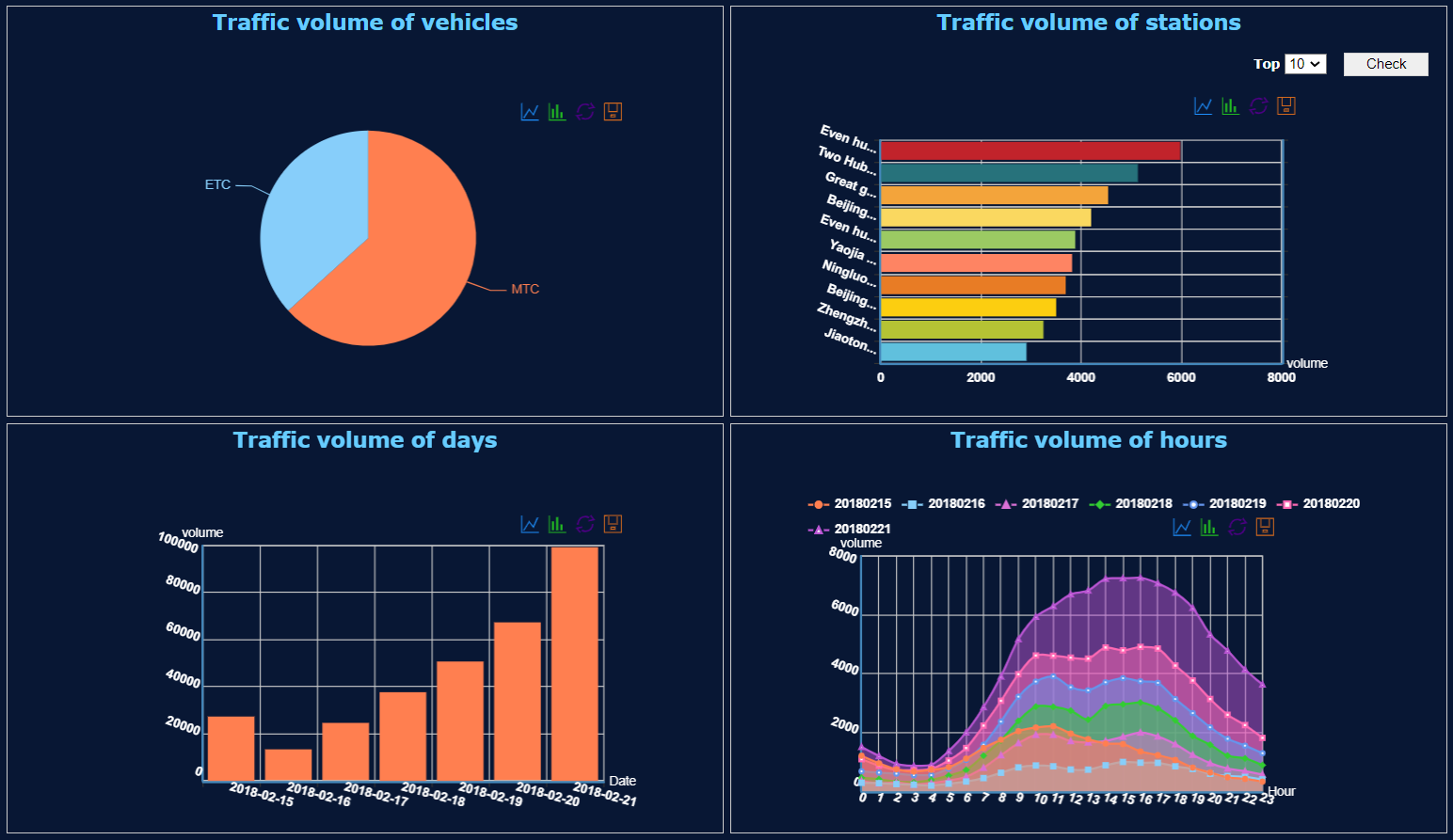}
		\caption{Highway situation predictive analysis on holidays in various perspectives}
		\label{holiday}
	\end{minipage}
	\begin{minipage}[t]{1\linewidth}
		\centering
		\includegraphics[width=3.5in]{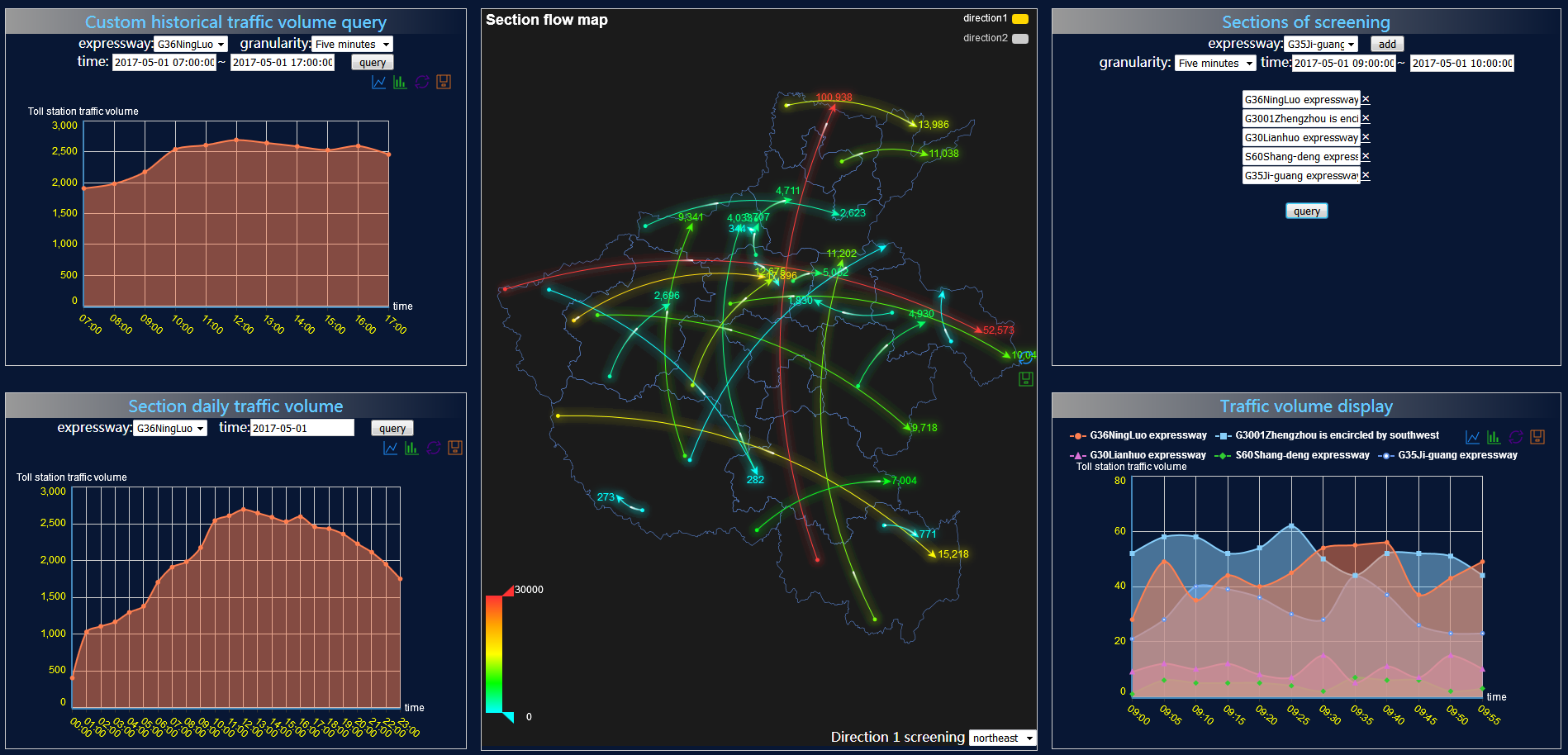}
		\caption{Highway traffic flow spatio-temporal correlation}
		\label{correlation}
	\end{minipage}
\end{figure}

On some of holidays like Spring Festival, toll-free policy would be carried out by Chinese government, and possible burst by travel of private cars makes much highway stress during those days. Through our ADGCRNN as basic prediction service, a situation analysis application on Spring Festival is implemented in system for business officers. As figure \ref{holiday}, on this 7-day holiday, the predictive traffic flow are represented in four perspectives: vehicular type proportion, toll station ranking, daily comparison, and hourly comparison on each date. 
In the perspective left-top, daily traffic flows are summarized from all the stations on seven days, and then divided into two types by driver identity (i.e., either MTC or ETC). In the perspective right-top, toll stations are ordered by summary of traffic flows on seven days. Spatial characteristics are ranked by a bar chart, and top-10 stations are presented. In the perspective left-bottom, respective dates with the summary of network-wide traffic flows are compared. Temporal characteristics are reflected in a histogram. In the perspective right-bottom, hours are compared on each of the seven days. Here,  fine-granularity traffic peaks are prominently high than others on the last two days, because return flows would bust intensively when a holiday is close to the end. All those perspectives are helpful for officers to mange highway transportation accordingly, and prove our work's extensive feasibility.

From the network-wide monitor in Henan highway, as figure \ref{correlation}, the spatio-temporal correlation employed in our ADGCRNN can be presented. Hundreds of toll stations compose the graph in our model, and only the start and end ones of expressway-lines are dotted on the map in the middle for concise visualization. In the left-top perspective, the traffic flows of expressway-line (i.e., the sum of station traffic flows in that line) are organized by customized query at various temporal resolutions. Such resolutions include five-minute, 15-minute, 30-minute, one-day, one-week and one-month. The left-bottom line chart outputs summarized traffic flows of toll stations at given prefecture-level cities. Officers can get further comparisons of expressway-line traffic flow in the same duration from the right perspectives. 
From such monitor perspectives of traffic flow, some interesting facts can be found further. For example, the node \emph{Zhengzhou South}, as the bustiest station in Henan highway, owns its traffic flow at week resolution the most related with that of \emph{East 3rd Ring} station 9 kilometers from the east; but its trend at daily resolution is the most affected by \emph{Wenhua Road} station 35 kilometers northern away.   
Accordingly, traffic flow can be demonstrated from multiple perspectives, and its spatio-temporal correlation even at various resolutions is proved complex and valuable in practice. 

\section{Conclusion}
In this paper, ADGCRNN model is proposed for traffic flow prediction in highway transportation. 
The spatio-temporal relationships of traffic flow at three resolutions are integrated by self-attention mechanism.
A novel dynamic graph cell is to obtain consistent spatio-temporal features from the combination of RNN and GCN, where multi-dynamic graphs, dynamic graph weights, and gated kernel are proved their effectiveness.
Extensive experiments on public datasets prove that our work can reduce MAE at least 3.85\% than state-of-the-art baselines, and each module of model effectively improves predictive results. Case studies in a real project show convincing benefits in practice.


\bibliographystyle{ACM-Reference-Format}
\bibliography{my_reference}

\appendix

\end{document}